%% file: root.tex
\documentclass[letterpaper, 10 pt, conference]{ieeeconf}  

\IEEEoverridecommandlockouts                              

\usepackage{graphics} 
\usepackage{epsfig} 
\usepackage{mathptmx} 
\usepackage{times} 
\usepackage{amsmath} 
\usepackage{amssymb}  
\usepackage{booktabs}
\usepackage{multirow}

\title{\LARGE \bf
TP3M: Transformer-based Pseudo 3D Image Matching with Reference Image
}

\author{Liming Han$^{1,2}$  Zhaoxiang Liu$^{1,2}$$^{*}$ Shiguo Lian$^{1,2}$$^{*}$
\thanks{$^{1}$AI Innovation Center, China Unicom, Beijing 100013, China.\newline
        $^{2}$Unicom Digital Technology, China Unicom, Beijing 100013, China.
        {\tt\small hanlm21, liuzx178, liansg@chinaunicom.cn \newline}
{This work is supported by the Funding of Beijing Association of Science and Technology Outstanding Engineer Growth Plan}}%
\thanks{$^{*}$Corresponding author}%
}

\begin{document}

\maketitle
\thispagestyle{empty}
\pagestyle{empty}

\begin{abstract}

Image matching is still challenging in such scenes with large viewpoints or illumination changes or with low textures. In this paper, we propose a Transformer-based pseudo 3D image matching method. It upgrades the 2D features extracted from the source image to 3D features with the help of a reference image and matches to the 2D features extracted from the destination image by the coarse-to-fine 3D matching. Our key discovery is that by introducing the reference image, the source image's fine points are screened and furtherly their feature descriptors are enriched from 2D to 3D, which improves the match performance with the destination image. Experimental results on multiple datasets show that the proposed method achieves the state-of-the-art on the tasks of homography estimation, pose estimation and visual localization especially in challenging scenes.

\end{abstract}

\section{INTRODUCTION}

Image matching, as a basic task in computer vision, finds corresponding points between two or more views of a scene. For example, it is an important module of Structure from Motion (SfM) \cite{lindenberger2021pixsfm, schonberger2016sfm}, Simultaneous Location and Mapping(SLAM) \cite{chen2021panoramicSLAM, cadena2016slam}, and visual localization \cite{sarlin2019hloc, taira2018inloc, yoon2021localization}. 


Detector-based image matching methods \cite{revaud2019r2d2, dusmanu2019d2net, yang2020ur2kid} often fail to get robust matching in challenging real-world image pairs due to the changes of illumination, texture, viewpoint, occlusion, blur, etc.. Detector-free image matching methods, such as LoFTR \cite{sun2021loftr} and MatchFormer \cite{wang2022matchformer}, extract features even from images with few textures and have achieved state-of-the-art results. 
Anyway, for both of above methods, to extract robust features is important for correct image matching, while that is often challenging when extracting features from only 2D image. 


\input{figs/Fig1}

However, the pseudo LiDARs \cite{qian2020pseudolidar, ma2020pseudolidar, weng2019pseudolidar} prove that the deep model can learn 3D information from multiple 2D images.  And, recently, Vision Transformer (ViT) \cite{dosovitskiy2020vit, zhao2022monovit} has achieved good performances in image matching task \cite{wang2022matchformer} and point cloud registration task \cite{guo2021pct, engel2021point_transformer}.


Inspired by pseudo LiDAR and ViT, we try to improve image matching by extending 2D features and 2D matching to 3D ones respectively with ViT. In contrast to pseudo LiDAR, the incorporation of reference images is aimed at facilitating the extraction of 3D features from the source image, rather than reconstructing depth or 3D maps in a manner akin to the source image. Generally, the reference image may be a different view of the source image and it is available in various tasks. Here, for the proposed method, we name it pseudo 3D. Additionally, based on 2D matching, the coarse-to-fine 3D matching is designed to find the matches between the source image's 3D features and destination image's 2D features. Here, both feature extraction and feature matching are constructed on ViT with respect to ViT's superiority mentioned above. As shown in Fig. \ref{fig:result}, when there are strong illumination changes and large viewpoint changes, our method TP3M obtains large number of correct matches while MatchFormer gets many mismatches. 

In summary, the paper's contributions include:

(1) TP3M is an end-to-end network. In contrast to pure 2D feature extraction and matching approaches, our method refines and enriches the features to 3D with the aid of reference image, and thus improves the matching performance. The experimental results are given to show that it achieves state-of-the-art performances on relative pose estimation and visual localization on multiple datasets.

(2) We design a pseudo 3D feature extraction method composed of 2D edge feature detection and 3D feature fusion. It extracts the source image's fine features with the aid of reference image, which contains the semi-dense and precise features with edge-aware attentions, the geometric shapes and description information of the scene. These features are robust in challenging scenes that have been proved by experimental results.
 
(3) We present a coarse-to-fine 3D matching approach that combines the coarse matching based on 2D edge features with the fine matching based on 3D features. Both the coarse and fine matching modules are constructed on ViT, whose effectiveness are proved by experimental results.

\section{RELATED WORK}

\subsection{Transformer in 2D Matching}

To solve the problem that traditional feature matching methods are not robust under challenging conditions, many ViT based methods have been proposed.
LoFTR \cite{sun2021loftr} is a detector-free method, which uses self-attention layer and cross attention layer to obtain the feature descriptors of two images. 
The global receptive field provided by Transformer allows LoFTR to generate dense matching in low texture regions. 
MatchFormer \cite{wang2022matchformer} has a robust hierarchical Transformer encoder and a lightweight decoder. 
Inside each stage of the hierarchical encoder, it interleaves self-attention for feature extraction and cross-attention for feature matching. 
MatchFormer is a good solution in terms of efficiency, robustness and accuracy. 
OETR \cite{chen2022oetr} proposes a novel overlap estimation method conditioned on image pairs with Transformer to constrain local feature matching in the commonly visible region. 
It is plugged into local feature detection and matching pipeline to mitigate potential view angle or scale variance. 
SuperGlue \cite{sarlin2020superglue} introduces a flexible context aggregation mechanism based on attention, enabling it to reason about the underlying 3D scene and feature assignments jointly. 
It learns the matches between two sets of interest points with a graph neural network (GNN),  which is a general form of Transformers \cite{joshi2020transformers}. 
However, feature matching with 2D information is difficult to achieve good results in challenging scenes due to the lack of 3D information.

\subsection{Transformer in 3D Matching}

Similar to feature matching, self and cross attentions in Transformer extract and match features from point clouds in point cloud registration task. 
GeoTransformer \cite{qin2022geotransformer}  proposes a geometric Transformer that learns geometric feature for robust superpoint matching. 
It is composed of a geometric self-attention module for learning intra-point-cloud features and a feature-based cross-attention module for modeling inter-point-cloud consistency. 
Following Transformer \cite{vaswani2017attention}, DCT-v2 \cite{wang2019dcp} uses an attention based module combining pointer network to predict a soft matching between the point clouds. 
REGTR \cite{yew2022regtr} uses a network architecture consisting primarily of Transformer layers containing self and cross attentions, and it predicts the probability each point lies in the overlapping region and its corresponding position in the other point cloud. 
As can be seen, most of existing work construct 3D feature or 3D matching based on point cloud that may not be available in most scenes. Differently, in this paper, we focus on 3D feature extraction and matching based only on 2D images.

\input{figs/Fig2}

\subsection{Multimodal Transformer}

Transformer can effectively extract the features from images and point clouds, and fuse them in the feature layer. 
Some cross modal feature fusion methods achieve state-of-the-art performance in 3D target detection. 
TransFuser \cite{Chitta2022TransFuser} is a novel multi-modal fusion Transformer to incorporate global context and pairwise interactions into the feature extraction layers of different input modalities. 
It dynamically detects uninformative tokens and substitutes these tokens with projected and aggregated inter-modal features. 
TokenFusion \cite{wang2022TokenFusion} proposes a multimodal token fusion method and allows the Transformer to learn correlations among multimodal features, while the single-modal Transformer architecture remains largely intact. 
It surpasses state-of-the-art methods in three typical vision tasks: multimodal image-to-image translation, RGB-depth semantic segmentation, and 3D object detection with point cloud and images. 
TransFuseGrid \cite{salazar2022transfusegrid} is a Transformer-based, multi-scale fusion architecture to fuse multi-camera and LiDAR features and predict semantic grids. 
However, it is difficult to fuse the features of different sensors. 
Additionally, some work adopts Transformer to construct depth map or 3D map of the scene from multiview images, such as Neural Radiance Fields(NeRF) \cite{mildenhall2021nerf} or pseudo LiDAR \cite{qian2020pseudolidar}. They prove that it is available to reconstruct 3D information from multiple 2D images.
Inspired by Neural Radiance Fields(NeRF) \cite{mildenhall2021nerf} and pseudo LiDAR \cite{qian2020pseudolidar}, we extend the source image to image pairs by introducing a reference image, from which 3D features with better matching performance are extracted by Transformers. Note that, in our method, it is 3D feature to be constructed while not depth map or 3D map.

\section{METHOD}

As shown in Fig. \ref{fig:structure}, given the challenging image pair consisting of the source image $I_A$ and destination image $I_B$, TP3M estimates robust and accurate matches as follows. 
As a prerequisite, we introduce the reference image $I_C$ that is close to the viewpoint of $I_A$. First, the 2D edge features are extracted from each of the three images by a Transformer. Then, the 2D matches between the source image $I_A$ and reference image $I_C$ are computed by another Transformer, together with the source image's 2D edge features are fused to construct the 3D features of source image. Finally, the 3D matches between the source image $I_A$ and destination image $I_B$ are computed by the third Transformer according to their 2D matches, source image's 3D features and destination image's 2D features. 

\subsection{2D Edge Feature Detection}

\input{figs/Fig3}

Edge features contain stable texture and geometric information of images. 
Considering that multi-scale feature fusion can effectively extract edge and descriptor, we use a 3-layers network to extract pyramid features $F_A$, as shown in Fig. \ref{fig:2D_feature}. Generally, the position embedding method in ViT cannot directly obtain low-level feature information, which limits local feature matching. Here, the edge feature is improved on the basis of Positional Patch Embedding (PE) of MatchFormer to solves this problem. 
In MatchFormer, Positional PE enhances the position information of the patch and extracts denser features by increasing the depthwise convolution, and after that a self-attention in Transformer is usually used for feature extraction of the image itself. However, these uniformly distributed dense features detected by MatchFormer are lack of significance judgment. 
Different from MatchFormer, after the Positional PE and self attention, we use the BiMLA proposed in EDTER \cite{Pu_2022_EDTER} to calculate the gradient of the feature points. In detail, in our method, the BiMLA is applied to the first two scales in the pyramid instead of global and local processing separately. As a result, our BiMLA is of only 3-layers, and thus more lightweight. 
\label{subsec:2D Edge Feature Detection}

\subsection{2D Feature Matching}
After extracting the multi-scale edge features of the image, a  cross-attention in Transformer framework is used to process local feature matching layer by layer. 
Specifically, for $f_{A3}$ and $f_{B3}$, the top-level features of the pyramid corresponding to images $I_A$ and $I_B$, their cross-attention is calculated, and we get the confidence matrix $P_3$. 
Following LoFTR, we select the matches with confidence higher than a threshold $\theta_3$, and we further enforce the mutual nearest neighbor (MNN) criteria, which filters possible outlier coarse matches. 
When the number of matches on the 3rd layer is higher than a given threshold $N_3$, the confidence matrix $P_2$ for the 2nd layer is calculated with the features $f_{A2}$ and $f_{B2}$. 
Similarly, if the number of matches on the 2nd layer is also higher than the threshold $N_2$, we calculate the confidence matrix $P_{\rm{2D-2D}}$ corresponding to features of the original image. 
Gathering all the matches produces the final 2D feature matching result.

Once the number of matches on a certain layer is lower than the threshold, 2D feature matching stops on the layer and fails. 
Generally, there are two failing cases: 1) The overlap of the scenes between $I_A$ and $I_B$ is small and it obtains few matches; 2) The overlap of the scenes is large enough, but there are great challenges in terms of large viewpoint, lighting changes and lack of texture. 
Usually, a large number of matches is produced from $I_A$ and $I_C$ under similar conditions. 
However it produces fewer matches for $I_A$ and $I_B$ in challenging scenes. Considering that this feature matching processes only 2D features, we also call it 2D-2D matching for simplicity.
\label{subsec:2D Feature Matching}

\subsection{Pseudo 3D Feature Extraction}

Inspired by PETR \cite{liu2022petr}, VPFusion \cite{mahmud2022vpfusion} and multi-view 3D reconstruction with Transformers \cite{wang2022mvster, ding2022transmvsnet}, we capture 3D-structure-aware context and pixel-aligned image features. 
After 2D feature matching between $I_A$ and $I_C$ as mentioned in Section \ref{subsec:2D Feature Matching}, we obtain the position features $P_{\rm{2D-2D}}(A,C)$, which represent the spatial relationship of image features. Following PETR, we use a network to transform the position features. Differently, the fully connected (FC) layer at the end of PETR network is removed in order to make the dimension of 3D position features consistent with that of 2D features. Then, we fuse the transformed 3D position features with the corresponding 2D features $F_A$ by addition operator to get the fusion features $F_{\rm{3D}}$.
Thus, $F_{\rm{3D}}$ contains both the image's 2D description information and its 3D spatial geometric information.
\label{subsec:Pseudo 3D Feature Extraction}

\subsection{Pseudo 3D Matching}
For challenging scenes with large light or viewpoint changes, 2D feature matching usually gets few matches and even some mismatches. To get better matches, we propose the corarse-to-fine matching scheme composed of two steps: the 2D feature matching as coarse matching and 3D feature matching as fine matching.
Inspired by BEV methods \cite{bai2022transfusion, chen2022persformer, liu2022petrv2}, they use 2D image features as K, V, and 3D features as Q. 
We introduce a cross-attention layer
to 3D feature matching, where $P_{\rm{2D-2D}}(A,B)$ is similar to image guidance \cite{bai2022transfusion}. The confidence matrix $P_{\rm{3D}}$ measures the similarity of 2D and 3D features.  
Due to the information of coarse matches, fine matching is easier to converge and obtain the optimal solution. 
According to the continuity of geometric features, we set the sliding window size. When the probability is large enough for consecutive points in the window, these features are successfully matched.

In fact, coarse matches are calculated by the cross-correlation with the statistical and distributional information of the image pixels. 
When the lighting or viewpoint changes, the pixels of the image change greatly, and thus coarse matches only establish correct matches between a few distinct features. 
Differently, position information is additionally introduced to form 3D features that contain geometric information in addition to image statistical information. 
Thus, even if the image pixels change greatly, better results will be obtained due to the invariance of geometric features. 
Additionally, for low-texture scenes, more matches are also established at edge points in the image due to using edge-aware features.
Considering that this feature matching processes both 2D features and 3D features, we also call it 2D-3D matching for simplicity.
\label{subsec:Pseudo 3D Matching}

\subsection{Supervision and Training}
In TP3M, the 2D edge feature detection network is trained separately, while the 2D matching, 3D feature extraction and 3D matching networks are trained together. For the latter, the loss consists of 2D feature matching loss $L_{\rm{2D-2D}}$, 3D feature matching loss $L_{\rm{2D-3D}}$ and 3D features loss $L_{\rm{3D}}$. That is
\begin{equation}
  L=L_{\rm{2D-2D}}+L_{\rm{2D-3D}}+L_{\rm{3D}}.
  \label{eq:Loss_sum}
\end{equation}

For edge features, the edges detected by Canny \cite{canny1986} are taken as ground truth. We compute the confidence matrix and SfM results with the camera pose, RGB and depth maps during training of TP3M. The confidence matrix is defined as the mutual nearest neighbor of the reprojection distance of the two sets of images. The SfM results are taken as the ground truth of $L_{\rm{3D}}$, and the confidence matrix as the ground truth of $L_{\rm{2D-2D}}$ and $L_{\rm{2D-3D}}$. We calculate them as 
\begin{equation}
  L_{\rm{2D-2D}}=-{\frac{1}{\lvert{M_{\rm{2D}}^{gt}}\rvert} } \sum\limits_{(i,j)\in{M_{\rm{2D}}^{gt}}} \alpha_{ij}\log_{}{P_{\rm{2D-2D}}(i,j)}
  \label{eq:Loss_2D2D},
\end{equation}
\begin{equation}
  L_{\rm{2D-3D}}=-{\frac{1}{\lvert{M_{\rm{3D}}^{gt}}\rvert} } \sum\limits_{(i,j)\in{M_{\rm{3D}}^{gt}}} \beta_{ij}\log_{}{P_{\rm{2D-3D}}(i,j)}
  \label{eq:Loss_2D3D},
\end{equation}
\begin{equation}
  L_{\rm{3D}}=-{\frac{1}{\lvert{M_{\rm{3D}}^{gt}}\rvert} } \sum\limits_{(i,j)\in{M_{\rm{3D}}^{gt}}} \gamma_{ij}\log_{}{D_{\rm{3D}}(i,j)}
  \label{eq:Loss_3D}.
\end{equation}
Here, $M_{\rm{2D}}^{gt}$ denote the edges in the image, and $M_{\rm{3D}}^{gt}$ the edges in SfM reconstructed results with ground truth information. 
Compared to $M_{\rm{2D}}^{gt}$, $M_{\rm{3D}}^{gt}$ is not affected by scale and occlusion. 
$\alpha_{ij}$, $\beta_{ij}$ and $\gamma_{ij}$ are the significance weights of edge points. 
They are calculated according to Laplacian operator. 
$P_{\rm{2D-2D}}(i,j)$ is the matching probability of 2D edge features, $P_{\rm{2D-3D}}(i,j)$ is the matching probability of 2D and 3D features. 
$D_{\rm{3D}}(i,j)$ is a european distance probability between the estimated position and ground truth in the set where $P_{\rm{2D-3D}}(i,j)$ is higher than a threshold. 

When using dual softmax for matching, the matching probabilities are calculated as 
\begin{equation}
  \begin{aligned}
  P_{\rm{2D-2D}}(i,j)=&{\rm softmax}(S_{\rm{2D-2D}}(i,\cdot))_j\cdot\\
    &{{\rm softmax}(S_{\rm{2D-2D}}(\cdot,j))_i},
  \end{aligned}
  \label{eq:P_2D2D}
\end{equation}
\begin{equation}
  S_{\rm{2D-2D}}(i,j)=\frac{1}{\omega}\left\langle {F_A(i)},{F_B(j)} \right\rangle,
  \label{eq:S_2D2D}
\end{equation}
\begin{equation}
  \begin{aligned}
  P_{\rm{2D-3D}}(i,j)=&{\rm softmax}(S_{\rm{2D-3D}}(i,\cdot))_j\cdot\\
    &{{\rm softmax}(S_{\rm{2D-3D}}(\cdot,j))_i},
  \end{aligned}
  \label{eq:P_2D3D}
\end{equation}
\begin{equation}
  S_{\rm{2D-3D}}(i,j)=\frac{1}{\omega}\left\langle {F_{\rm{3D}}(i)},{F_B(j)} \right\rangle,
  \label{eq:S_2D3D}
\end{equation}
\begin{equation}
\begin{aligned}
  D_{\rm{3D}}(i,j)=\frac{1}{\delta(i)}{\Vert{j-j^{gt}}\Vert}_2.
  \label{eq:D_3D}
\end{aligned}
\end{equation}
Here, $F_A$ and $F_B$ are the 2D edge features of the images with self-attention, $F_{\rm{3D}}(i)$ is the 3D features, ${\Vert{j-j^{gt}}\Vert}_2$ is the distance between the estimated position and the ground true of the matching point in SfM results, and $\delta(i)$ the feature weight calculated according to the confidence matrix.

We train the models on Scannet \cite{dai2017scannet} and MegaDepth \cite{li2018megadepth} respectively. 
On Scannet, we select 2.3 million groups as the training set and 1500 groups as the test set. 
The models are trained using Adam \cite{kingma2014adam} with initial learning rate $1\times10^{-4}$ and batch size 64. 
On MegaDepth, we select 30000 groups for training. Same as MatchFormer and LoFTR, we use 1500 groups for testing. 
The models are trained using Adam with initial learning rate $1\times10^{-5}$  and batch size 16. 
Maintain the same experimental conditions as MatchFormer, all models are trained on 64 NVIDIA A100 GPUs, and tested on 8 NVIDIA A100 GPUs.
Although the introduction of reference images increases computational overhead, our method demonstrates a comparable computational cost to MatchFromer during testing, as we only calculate edge features.
\label{subsec:Supervision and Training}

\label{sec:method}

\section{EXPERIMENTS}

\subsection{Homography Estimation} 



We evaluate the impact of matching results on computing homography matrices on the HPatches \cite{balntas2017hpatches} benchmark which has significant illumination changes and large changes in viewpoint. The homography is estimated with the RANSAC method and is compared with the ground-truth. The area under the cumulative curve (AUC) is reported on threshold values of 1, 3 and 5 pixels, respectively. We use the default hyperparameters in the original implementations for all the baselines.

As shown in Tab. \ref{tab:Homography}, TP3M outperforms the existing baselines in homography experiments. 
For MatchFormer, although numerous matches were established in certain scenarios, no significant improvement was observed in terms of the AUC. TP3M obtains more correct matches than MatchFormer in the scenes with large viewpoint and illumination changes, and achieves the best performance on HPatches compared to SuperGlue+SuperPoint. This is attributed to the robust features, such as corners and edges.
\input{tables/tab1}
\label{subsec:Homography Estimation}

\subsection{Indoor Pose Estimation}



Indoor pose estimation is very challenging due to low textures, high self-similarity and complex spatial structures. 
We utilize the challenging indoor dataset ScanNet \cite{dai2017scannet} to demonstrate whether TP3M is able to learn 3D features from images to overcome these challenges.

Following SuperGlue, we report the pose error AUC at (5°, 10°, 20°) thresholds, where the pose error is the maximum of the angular errors in rotation and translation. 
The fundamental matrix is calculated by RANSAC method with matches. 
Then we get the relative pose and use the epipolar distance to calculate the accuracy P of matching results. 
As shown in Tab. \ref{tab:Indoor}, we report the AUC of the pose error in percentage and the matching precision (P) at the threshold of $5\times10^{-4}$.

\input{tables/tab2}

TP3M achieves the best performances compared to other methods. It extracts the corresponding 3D features from source and reference images, the number of matches is still sufficient for challenging indoor scenes with high accuracy. Firstly, the edge information is included by 3D features, which improves the invariance against changes. Secondly, for indoor complex spatial structure and repeated similarity, the 3D features can pay attention to global matching well and improve the matching accuracy.
\label{subsec:Indoor Pose Estimation}

\subsection{Outdoor Pose Estimation}




Due to the influence of illumination and seasonal changes in outdoor data, image matching is challenging. We choose outdoor dataset MegaDepth \cite{li2018megadepth} to evaluate our method with baselines, and adopt the the same metrics on pose error AUC as the indoor pose estimation task. The matching precision (P) is reported at the threshold of $1\times10^{-4}$ .

For outdoor dataset, the appearance of the scene is significantly different in the local region due to drastic lighting changes, which leads to a serious drop in the success rate of feature matching for detector-based methods. 
SuperGlue+SuperPoint \cite{detone2018superpoint} cannot work for many image pairs. 
While the detector-free methods obtain more matches, and the correct matches appear in the centralized distribution of a region in outdoor challenging scenes. 
There are some mismatches in other regions. 
The increase in the number of matches does not contribute to improving the pose accuracy. 
We also find that, on the MegaDepth dataset, there are some mismatches in the lower right part of the sample image in MatchFormer method, while the accurate matches are in a wider range in TP3M method. 
Compared with other methods, the edge-aware features of TP3M are relatively less affected by lighting, especially 3D features can use global geometric matching to suppress local mismatches, so TP3M performs well in outdoor evaluation as shown in Tab. \ref{tab:Outdoor}.

\input{tables/tab3}
\label{subsec:Outdoor Pose Estimation}

\subsection{Visual Localization}
\input{tables/tab4}
\input{tables/tab5}



Robust image matching is very helpful for visual localization \cite{zhang2021long_term, toft2020long_term}. So we evaluate our method on Aachen Day-Night dataset \cite{sattler2018aachen, sattler2012aachen} and InLoc \cite{taira2018inloc} dataset. We construct the scene's 3D models through SfM by use of various feature extraction and matching methods, including LoFTR, MatchFormer, SuperGlue+SuperPoint, ASpanFormer and ours. Based on the model, the absolute pose of the query image is computed through 2D-3D matching in Hloc \cite{sarlin2019hloc}. Finally the queried pose is compared with the ground truth.

Tab. \ref{tab:Aachen} and Tab. \ref{tab:InLoc} show the visual localization evaluation results on the outdoor dataset Aachen Day-Night and indoor dataset InLoc respectively. 
Here, MatchFormer and LoFTR is utilized as the feature matching module to complete the visual localization task along with the localization pipeline HLoc. 
The results show that TP3M obtains better performance in improving localization accuracy when using multiple image features.
Though MatchFormer extracts many features to establish a lot of matches in weak texture areas, see from Tab. \ref{tab:Aachen}, the accuracy of MatchFormer is not significantly improved compared with the accuracy of SuperGlue+SuperPoint. 
The features extracted by MatchFormer can only be matched in each pair of images, and they are not associated with more images. 
The number and description information of features in the 3D model is the key to achieve accurate 2D-3D matching. 
When the 3D model is established by HLoc with LoFTR and MatchFormer, it is difficult to capture enough essential geometric features and description information in the scene due to lack of data association. 
Differently, the edge-aware features elegantly establish multi-view feature associations, so the 3D model of TP3M contains more geometric information, and TP3M predicts more correct matches. 

\input{tables/tab6}
\input{figs/Fig8}
\label{subsec:Visual Localization}

\subsection{TP3M Structural Study}

\textbf{Ablation study}. 
To analyze the contributions of 2D edge features, coarse matching, fine matching and the number of reference images in the matching process, we design the ablation experiment as shown in Tab. \ref{tab:Ablation}. .

The results show that:
1) The accuracy decreases significantly when the 2D features detection removes the edge-aware module BiMLA. 
It indicates that the edge-aware module can accurately locate the position of edge features, inform the feature matching to focus more on the edges and thus improve the accuracy of pose estimation. 
2) When TP3M operates without fine matching, its accuracy markedly decreases, approaching levels comparable to LoFTR. 
It shows that 3D features can effectively represent scene geometric features and achieve robust feature matching when illumination and viewpoint changing, while 2D features cannot. 
3) The accuracy of pose estimation drops by 1.05 without coarse matching. 
It indicates that coarse matching can establish the initial value of fine matching and thus help to improve the accuracy. 
4) According to the metrics in Section \ref{subsec:Visual Localization}, 3D features are established between multiple reference and source images. 
3 references enhance the accuracy by +0.86 in the experiment. When adding more references, 3D features become richer and more accurate matching would be obtained.

\textbf{Visualizing Attention}. 
As shown in Fig. \ref{fig:Visualizing_attention}, we show the weights of 2D self attention, 3D self attention, 2D-2D cross attention and 2D-3D cross attention in Transformer, which are used for 2D edge feature detection, 2D matching and 3D matching. 
2D self attention focuses on the relationship between itself and other surrounding edge features. 
3D self attention removes the points which are too far away, and increases the relationship between itself and surrounding significant points. 
2D-2D cross attention determines feature matching relationship in a large range, while 2D-3D cross attention restricts the matching to a small correct range.
\label{subsec:TP3M Structural Study}

\label{sec:experiments}

\section{CONCLUSION}

We have presented a Transformer-based pseudo 3D image matching method called TP3M consisting of 3D feature extraction and 3D feature matching. In feature extraction, the source image's 2D feature is computed by Transformer and upgraded to 3D feature with the aid of a reference image. In feature matching, the source image's 3D feature is compared with the dest image's 2D feature in a coarse-to-fine manner. Experimental results on multiple datasets show that TP3M achieves the state-of-the-art in such tasks as homography estimation, relative pose estimation and visual localization. It proves that the proposed 3D feature contains more information invariant with changes than traditional 2D feature does. Thus, it is more suitable for challenging scenes such as visual mapping in life long SLAM with changeable lighting, seasons and viewpoints.



\bibliographystyle{IEEEtran}
\bibliography{IEEEabrv,references}


\end{document}

%% file: figs/Fig1.tex
\begin{figure}[t]
    \centering
        \includegraphics[width=\linewidth]{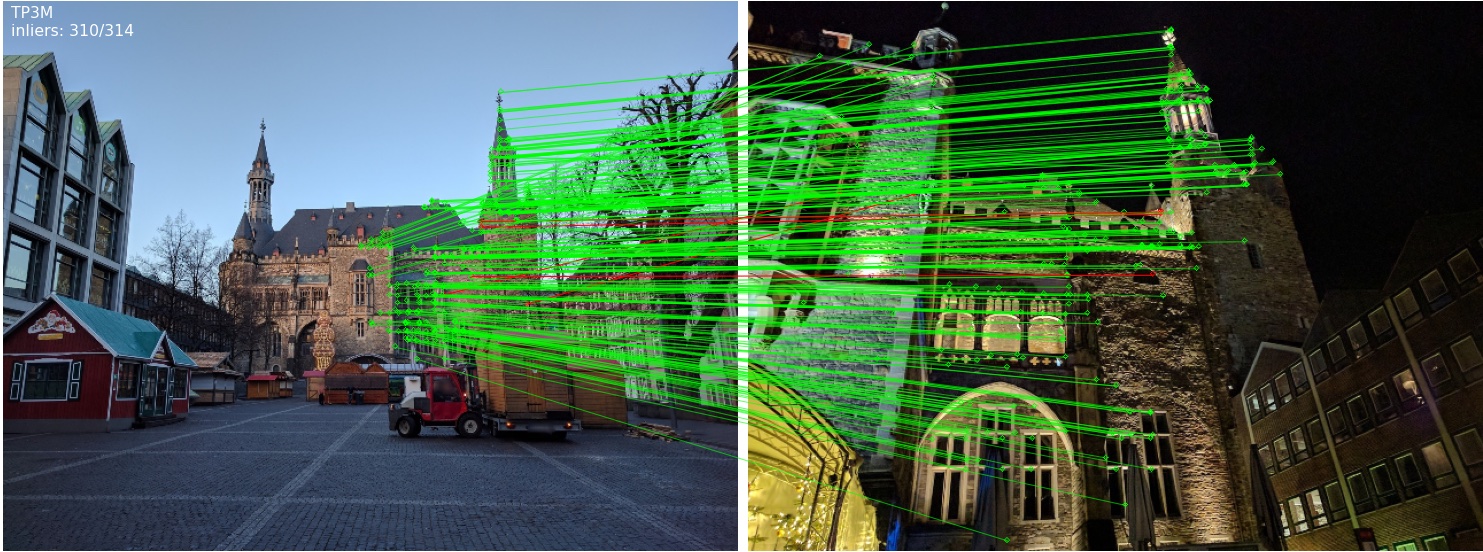} \\
        \includegraphics[width=\linewidth]{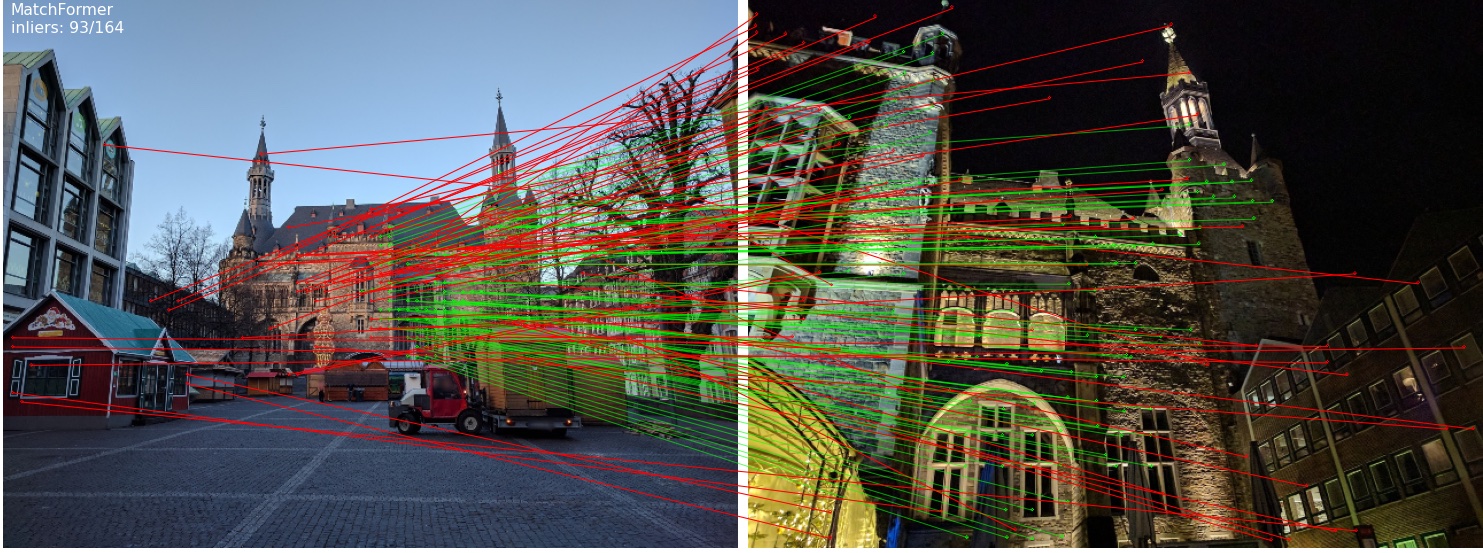}
    \caption{Comparison between TP3M and MatchFormer. Seen from the challenging image pairs with large viewpoint and illumination changes on the Aachen-Day-Night dataset, matching with TP3M results in more accurate poses than MatchFormer (correspondences colored by red lines represent epipolar error at 0.5m,5°).}
    \label{fig:result}
\end{figure}

%% file: figs/Fig2.tex
\begin{figure*}[t]
    \centering
        \includegraphics[width=0.9\linewidth]{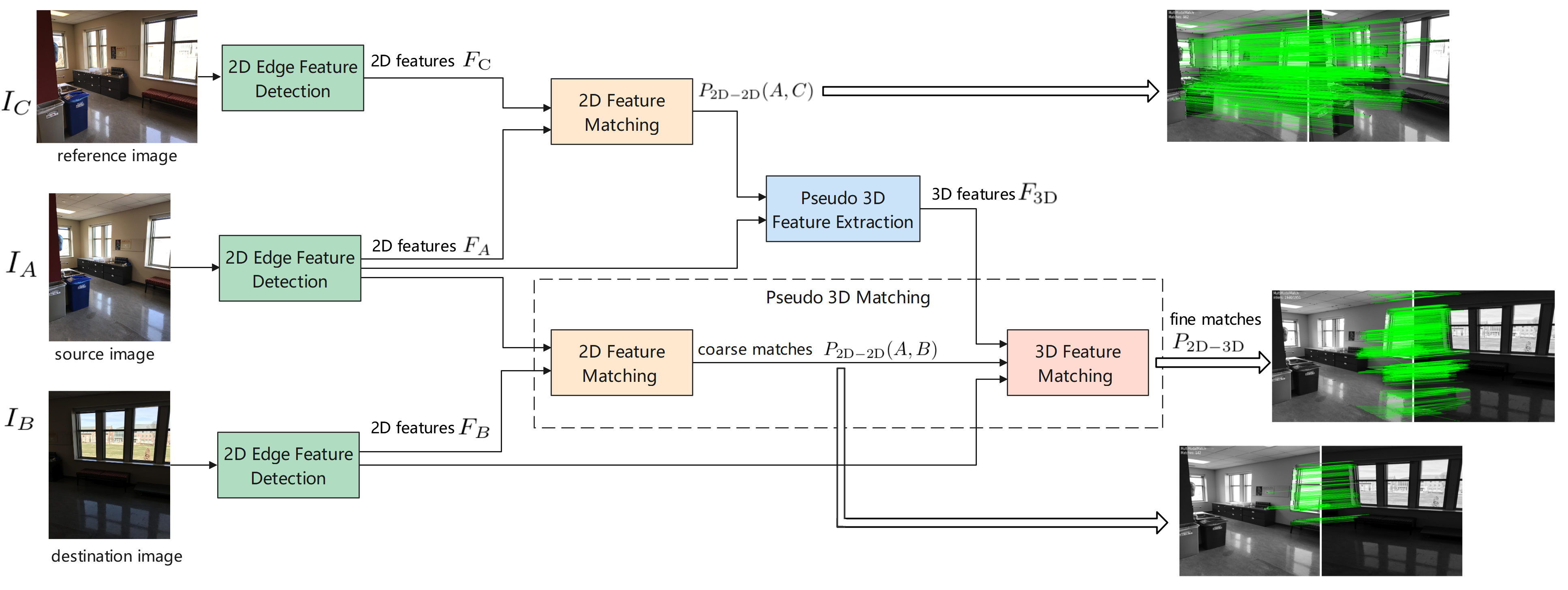} 
    \caption{Overview of the proposed TP3M. It includes four key modules: Transformer-based self-attention for 2D edge feature detection(\ref{subsec:2D Edge Feature Detection}); Transformer-based cross-attention for 2D feature matching (\ref{subsec:2D Feature Matching}); Pseudo 3D feature extraction (\ref{subsec:Pseudo 3D Feature Extraction});  Coarse-to-fine pseudo 3D matching between 2D features and 3D features(\ref{subsec:Pseudo 3D Matching}).}
    \label{fig:structure}
\end{figure*}

%% file: figs/Fig3.tex
\begin{figure}[t]
    \centering
        \includegraphics[width=\linewidth]{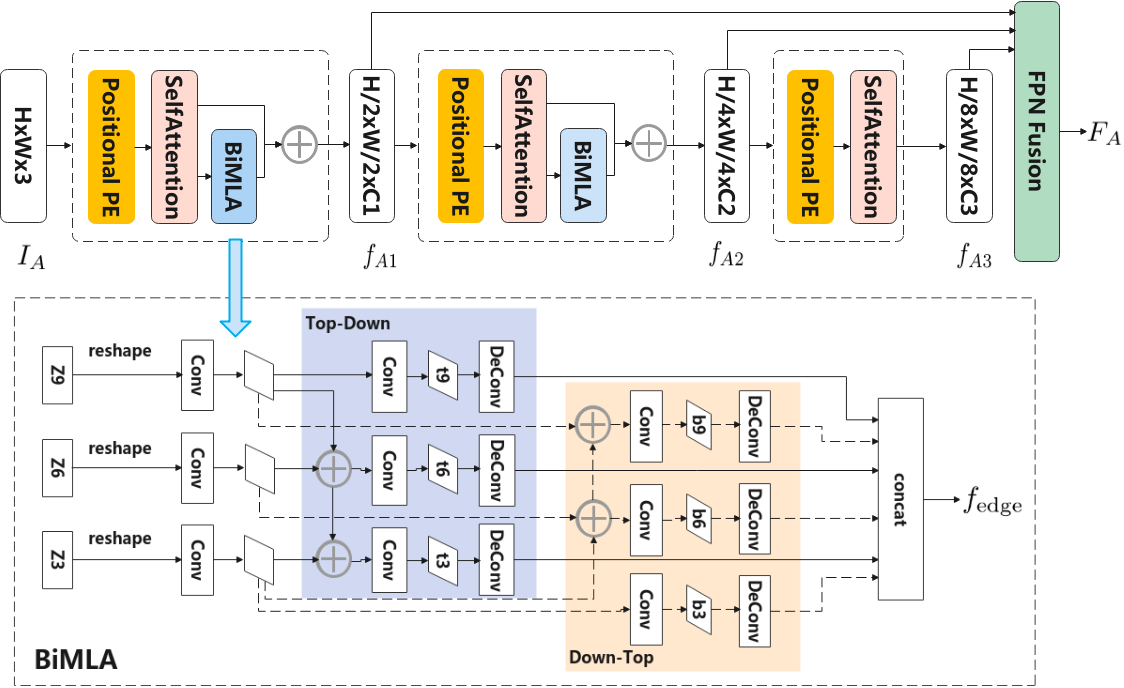} 
    \caption{2D edge feature detection with $I_A$ as example.}
    \label{fig:2D_feature}
\end{figure}

%% file: tables/tab1.tex
\begin{table}[t]
\centering

\begin{tabular}{lcccc}
\toprule
\multirow{2}{*}{Method} & \multicolumn{3}{c}{Homography estimation AUC} & \multirow{2}{*}{matches}\\    
	                      &@1px  &@3px  &@5px\\  
\midrule
SupeGlue+SP &	0.42 &	0.71 &	0.81 &	0.5K \\
LoFTR	    &   0.32 &	0.65 &	0.74 &	4.7K \\
MatchFormer &	0.37 &	0.68 &	0.78 &	\textbf{4.8K} \\
Ours       &\textbf{0.49} &\textbf{0.76} &\textbf{0.84} &	3.2K \\
\bottomrule
\end{tabular}

\caption{
Homography estimation on HPatches.
}
\label{tab:Homography}
\end{table}

%% file: tables/tab2.tex
\begin{table}[t]
\centering

\begin{tabular}{lcccc}
\toprule
\multirow{2}{*}{Method} & \multicolumn{3}{c}{Pose estimation AUC} & \multirow{2}{*}{P}\\    
	                      &@5°&@10°&@20°\\  
\midrule
SupeGlue+SP\cite{detone2018superpoint} &	16.16&	33.81&	51.84&	84.4 \\
LoFTR	    &   22.06&	40.80&	57.62&	87.9 \\
MatchFormer &	24.31&	43.90&	61.41&	89.5 \\
ASpanFormer\cite{chen2022aspanformer} & 25.60&	46.00&	63.30&	- \\
Ours       &\textbf{26.21} &\textbf{50.16} &\textbf{66.33} &\textbf{91.6} \\
\bottomrule
\end{tabular}

\caption{Indoor pose estimation on ScanNet.}
\label{tab:Indoor}
\end{table}

%% file: tables/tab3.tex
\begin{table}[t]
\centering

\begin{tabular}{lcccc}
\toprule
\multirow{2}{*}{Method} & \multicolumn{3}{c}{Pose estimation AUC} & \multirow{2}{*}{P}\\    
	                      &@5°&@10°&@20°\\  
\midrule
SupeGlue+SP &	42.18&	61.16&	75.95&	- \\
LoFTR	    &   52.80&	69.19&	81.18&	94.80 \\
MatchFormer &	52.91&	69.74&	82.00&	97.56 \\
ASpanFormer &   55.30&	71.50&	83.10&	- \\
Ours       &\textbf{57.22} &\textbf{74.53} &\textbf{85.81} &\textbf{98.99} \\
\bottomrule
\end{tabular}

\caption{Outdoor pose estimation on MegaDepth.}
\label{tab:Outdoor}
\end{table}

%% file: tables/tab4.tex
\begin{table}[t]
\centering

\begin{tabular}{lcc}
\toprule
\multirow{2}{*}{Method} &day &night \\    
	                &\multicolumn{2}{c}{(0.25m,2°) / (0.5m,5°) / (1m,10°)}\\  
\midrule
SupeGlue+SP &	89.8 / \textbf{96.1} / 99.4 &	77.0 / 90.6 / \textbf{100.0} \\
LoFTR	  &    88.7 / 95.6 / 99.0 &	78.5 / 90.6 / 99.0 \\
ASpanFormer &          89.4 / 95.6 / 99.0 &	77.5 / 91.6 / 99.5 \\
Ours       &	       \textbf{91.3} / 95.9 / \textbf{99.6} &	\textbf{83.1} / \textbf{93.3} / 99.6 \\
\bottomrule
\end{tabular}

\caption{Visual localization generated by HLoc on Aachen Day-Night benchmark.}
\label{tab:Aachen}
\end{table}

%% file: tables/tab5.tex
\begin{table}[t]
\centering

\begin{tabular}{lcc}
\toprule
\multirow{2}{*}{Method} &DUC1 &DUC2 \\    
	                &\multicolumn{2}{c}{(0.25m,10°) / (0.5m,10°) / (1m,10°)}\\  
\midrule
SupeGlue+SP & 49.0 / 68.7 / 80.8 & 53.4 / \textbf{77.1} / 82.4 \\
LoFTR	  &  47.5 / 72.2/ 84.8 & 54.2 / 74.8 / 85.5 \\
MatchFormer &  46.5 / 73.2 / 85.9 & 55.7 / 71.8 / 81.7 \\
ASpanFormer &       \textbf{51.5} / 73.7 / 86.4 & 55.0 / 74.0 / 81.7 \\
Ours       &	    48.8 / \textbf{74.2} / \textbf{88.6} & \textbf{56.6} / 76.1 / \textbf{86.3} \\
\bottomrule
\end{tabular}

\caption{Visual localization generated by HLoc on InLoc.}
\label{tab:InLoc}
\end{table}

%% file: tables/tab6.tex
\begin{table}[t]
\centering

\begin{tabular}{lccc}
\toprule
\multirow{2}{*}{Method} & \multicolumn{3}{c}{Pose estimation AUC}\\    
	                      &@5°&@10°&@20°\\  
\midrule
No BiMLA &	22.39 & 43.81 & 56.03 \\
No coarse matches & 25.16& 48.36 & 63.53 \\
No fine matches & 16.23 & 31.22 & 49.03 \\
More reference(3 images) &\textbf{27.07} &\textbf{52.05} &\textbf{67.18} \\
\bottomrule
\end{tabular}

\caption{Ablation study of TP3M on ScanNet.}
\label{tab:Ablation}
\end{table}

%% file: figs/Fig8.tex
\begin{figure}[t]
    \centering
        \includegraphics[width=0.31\linewidth]{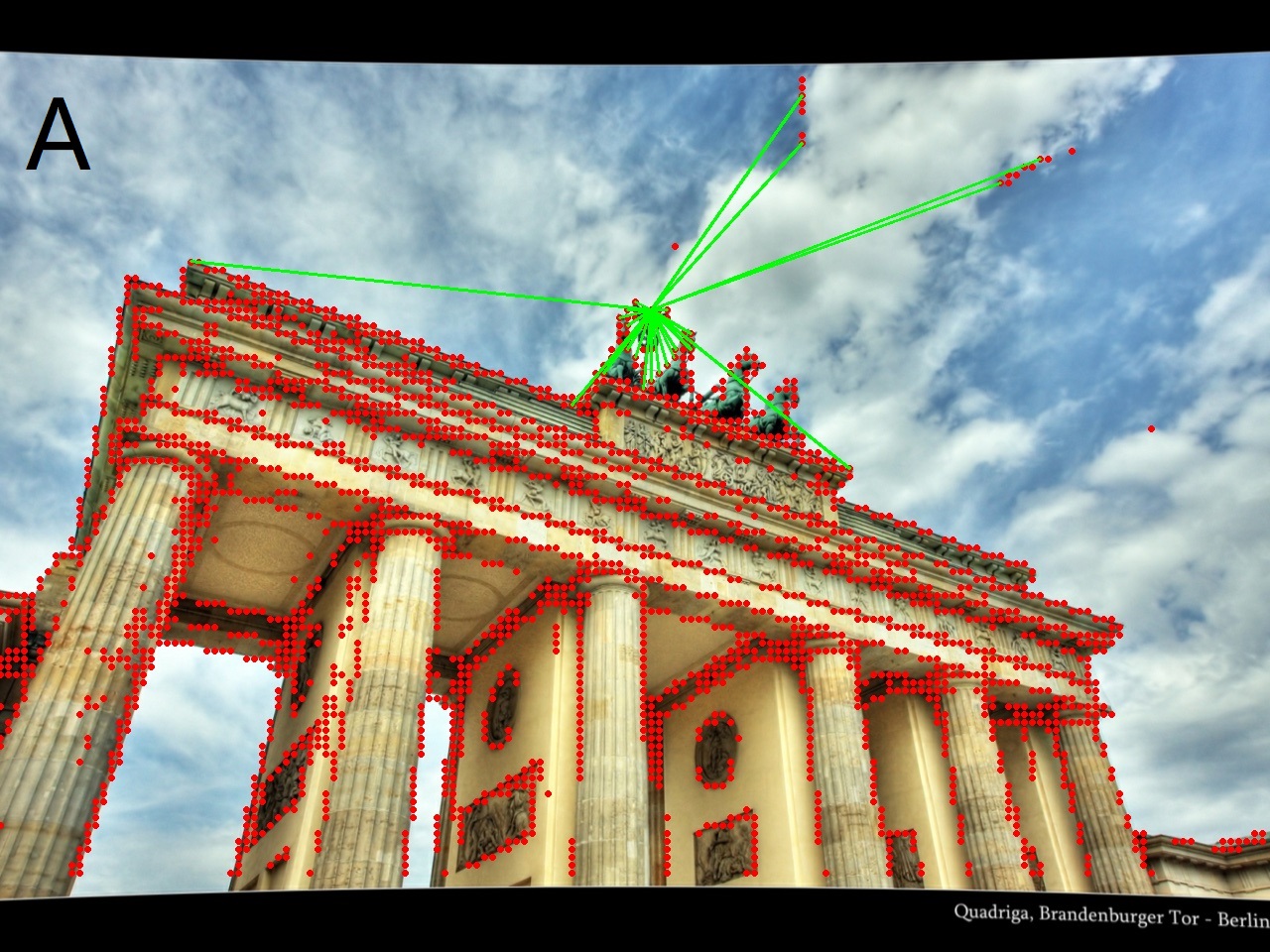} 
        \includegraphics[width=0.31\linewidth]{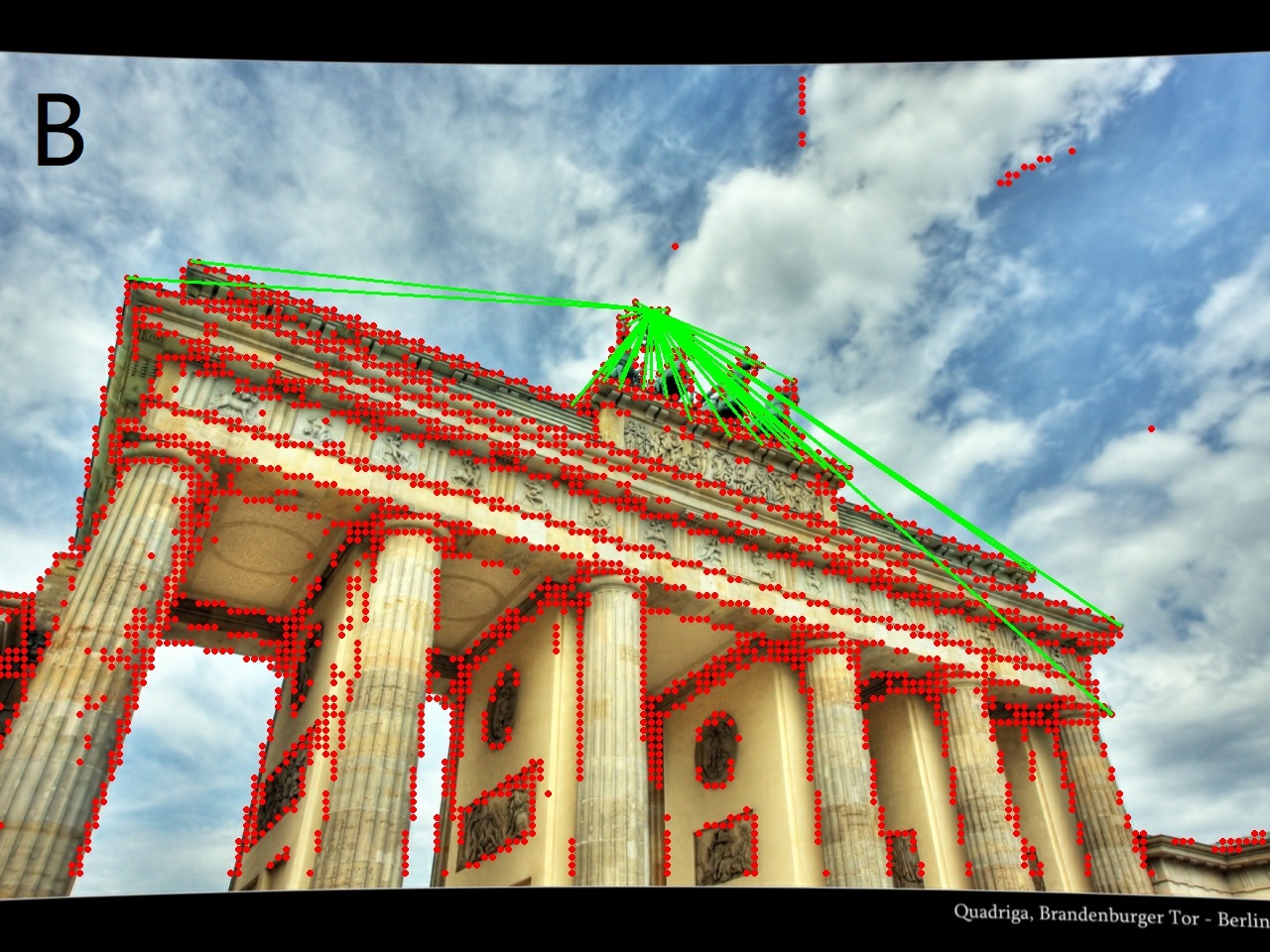}
        \includegraphics[width=0.31\linewidth]{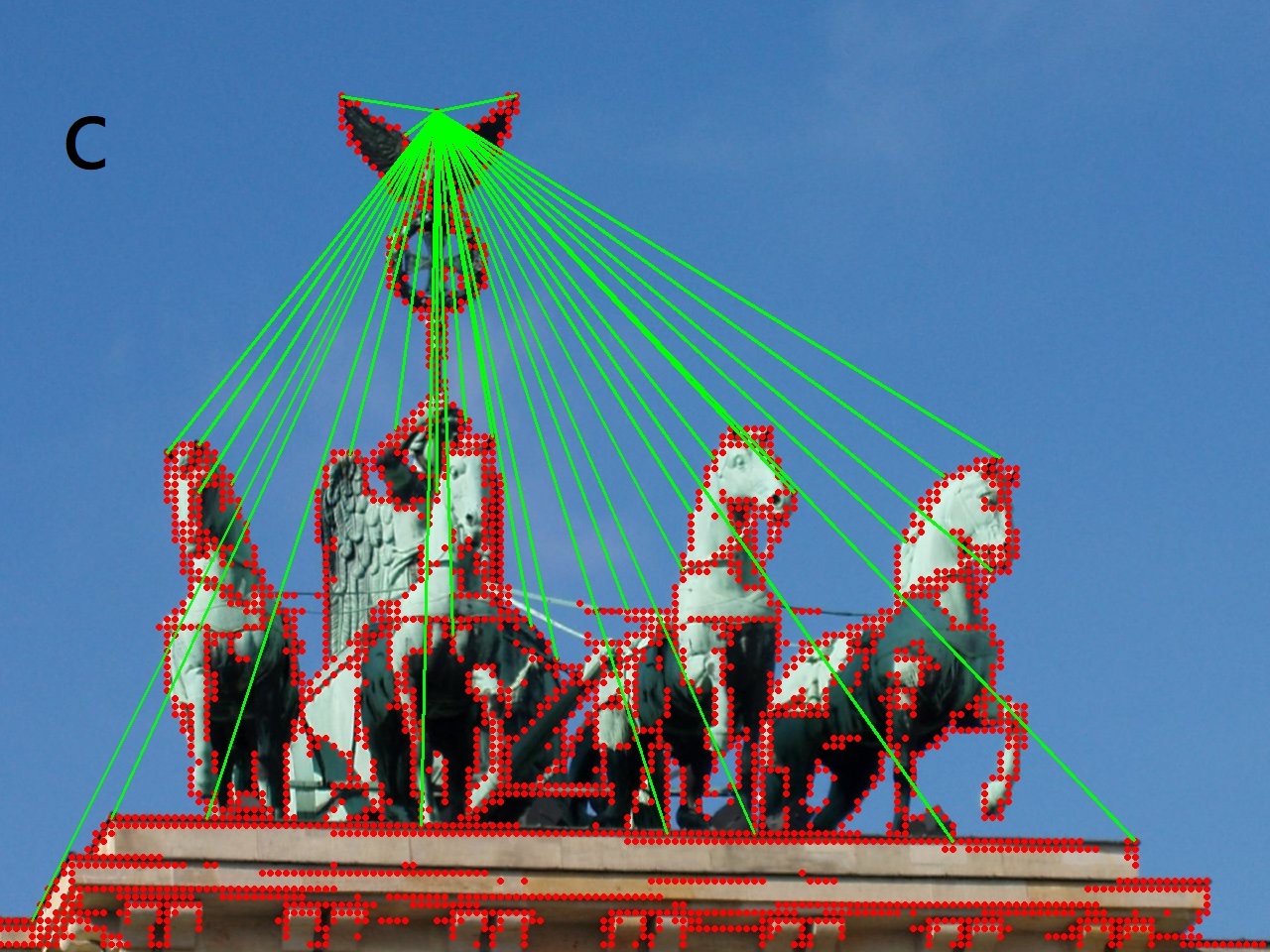} \\
        \includegraphics[width=0.47\linewidth]{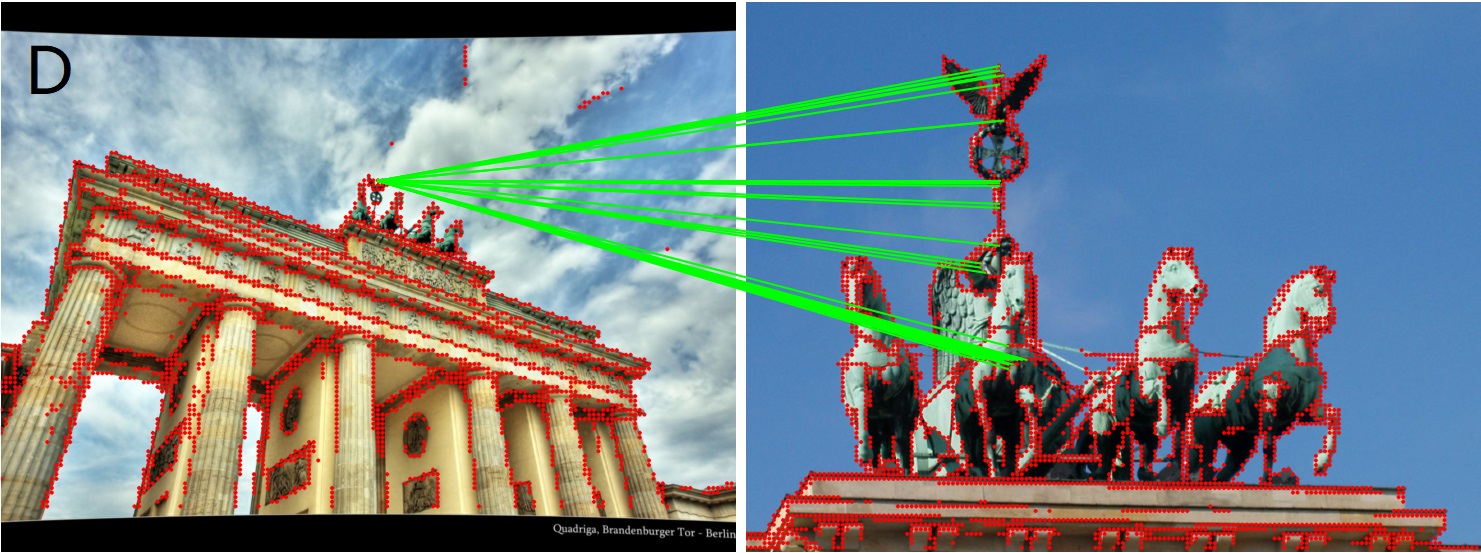}
        \includegraphics[width=0.47\linewidth]{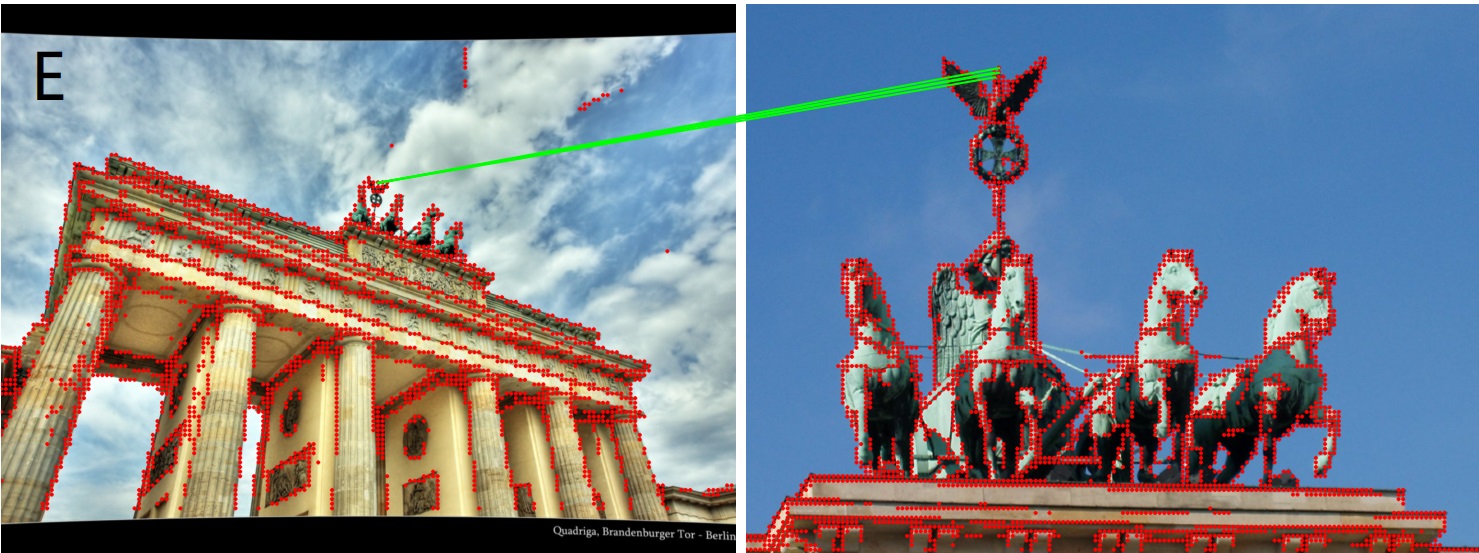}
    \caption{Visualizing attention. A: 2D self-attention in source image, B: 3D self-attention in source image, C: 2D self-attention in destination image. D : 2D-2D cross-attention, E : 2D-3D cross-attention between source and destination image.}
    \label{fig:Visualizing_attention}
\end{figure}